\documentclass[a4paper,fleqn]{cas-sc}

\usepackage[authoryear,longnamesfirst]{natbib}
\usepackage{algorithm}  
\usepackage{amsmath}
\usepackage{algpseudocode}
\usepackage{subcaption}
\usepackage{adjustbox}

\def\tsc#1{\csdef{#1}{\textsc{\lowercase{#1}}\xspace}}
\tsc{WGM}
\tsc{QE}
\tsc{EP}
\tsc{PMS}
\tsc{BEC}
\tsc{DE}

\begin{document}
\let\WriteBookmarks\relax
\def\floatpagepagefraction{1}
\def\textpagefraction{.001}

\shorttitle{2DMCG: 2D Mamba with Change Flow Guidance}

\shortauthors{JunYao Kuang et~al.}

\title [mode = title]{2DMCG: 2D Mamba with Change Flow Guidance for Change Detection in Remote Sensing}                      
\tnotemark[1]

\tnotetext[1]{This document is the results of the research
   project funded by the National Natural Science Foundation of China under Grants 52374155, Anhui Provincial Natural Science Foundation under Grant No. 2308085MF218, and PAPD of Jiangsu Higher Education Institutions.}


%
\author[1,2]{JunYao Kuang}[
                        style=chinese,
                        type=editor,
                        orcid=0009-0009-2779-772X]

\cormark[1]


\ead{6233112023@stu.jiangnan.edu.cn}



\affiliation[1]{organization={Engineering Research Center of Intelligent Technology for Healthcare, Ministry of Education, Jiangnan University},
    city={Wuxi Jiangsu},
    postcode={214122}, 
    country={China}}

\author[1,2]{HongWei Ge}[
                        style=chinese
                        ]
\cormark[2]


\affiliation[2]{organization={School of Artificial Intelligence and Computer Science, Jiangnan University},
    city={Wuxi Jiangsu},
    postcode={214122}, 
    country={China}}

\cortext[cor1]{Corresponding author}
\cortext[cor2]{Principal corresponding author}



\begin{abstract}
Remote sensing change detection (CD) has made significant advancements with the adoption of Convolutional Neural Networks (CNNs) and Transformers. While CNNs offer powerful feature extraction, they are constrained by receptive field limitations, and Transformers suffer from quadratic complexity when processing long sequences, restricting scalability. The Mamba architecture provides an appealing alternative, offering linear complexity and high parallelism. However, its inherent 1D processing structure causes a loss of spatial information in 2D vision tasks.
This paper addresses this limitation by proposing an efficient framework based on a Vision Mamba variant that enhances its ability to capture 2D spatial information while maintaining the linear complexity characteristic of Mamba. The framework employs a 2DMamba encoder to effectively learn global spatial contextual information from multi-temporal images. For feature fusion, we introduce a 2D scan-based, channel-parallel scanning strategy combined with a spatio-temporal feature fusion method, which adeptly captures both local and global change information, alleviating spatial discontinuity issues during fusion.
In the decoding stage, we present a feature change flow-based decoding method that improves the mapping of feature change information from low-resolution to high-resolution feature maps, mitigating feature shift and misalignment. Extensive experiments on benchmark datasets such as LEVIR-CD+ and WHU-CD demonstrate the superior performance of our framework compared to state-of-the-art methods, showcasing the potential of Vision Mamba for efficient and accurate remote sensing change detection.
\end{abstract}




\maketitle

\begin{figure}[t]
    \centering
    \includegraphics[width=\linewidth]{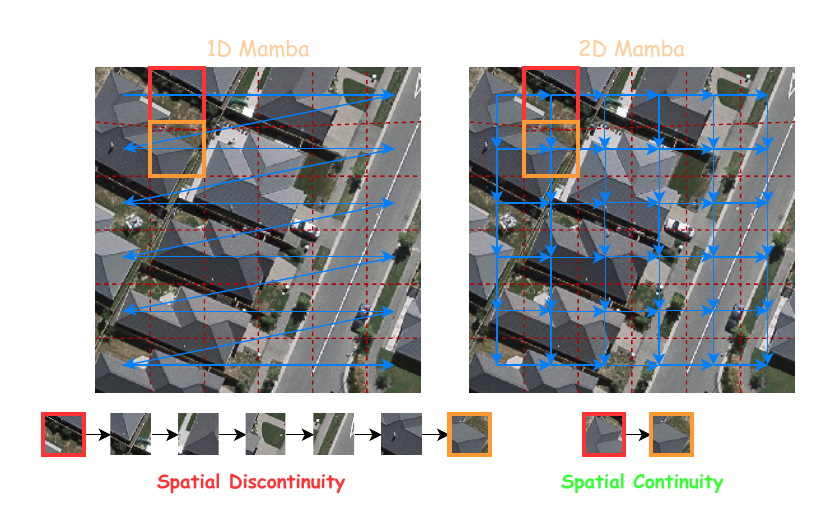}
    \caption{
        \textbf{Comparison of 1D and 2D Mamba-based methods.} 
        \textbf{Left:} 1D methods transform an image into a 1D sequence. This leads to \textit{spatial discontinuity} as adjacent patches (shown in \textcolor{red}{red} and \textcolor{orange}{orange}) become separated in the sequence.
        \textbf{Right:} 2D methods process the image in a 2D manner, maintaining \textit{spatial continuity}.
    }
    \label{fig:comparison}
\end{figure}

\section{Introduction}
Change Detection (CD) is a crucial task in remote sensing (RS) and Earth observation image analysis~\cite{daudt2018fully, chen2019change, fang2021snunet, zhang2020deeply, han2023hanet, Han2023CGNet, 9883686, Chen2022Remote, li2022transunetcd, zhang2023relation, chen2024changemamba}, with the goal of identifying changes on the Earth's surface by comparing co-registered images captured at different times~\cite{lu2004change}. The definition of "change" varies depending on the specific application, encompassing urban expansion, deforestation, vegetation changes, polar ice melting, and damage assessment~\cite{coppinp2004digital, coppin2002digital}. CD systems assign binary labels to each pixel to indicate whether a change has occurred between image acquisitions. This task is essential for generating maps that depict the evolution of land use~\cite{li2016superresolution}, urban coverage~\cite{wellmann2020remote}, and various other multi-temporal analyses. The study of CD has a long-standing history, evolving alongside advancements in image processing and computer vision techniques. Traditionally, information extraction from remote sensing images heavily relied on manual visual interpretation. However, automatic CD methods have gained increasing attention due to their potential to significantly reduce labor costs and time consumption~\cite{shi2020change, bai2023deep}.

In general, the core problem of Change Detection lies in how to effectively extract and identify differences from spectral, spatial, temporal, and multi-sensor data. The development of deep learning has introduced promising solutions to this problem, significantly improving both the accuracy and efficiency of CD tasks. Convolutional Neural Networks (CNNs) have become a popular choice for image analysis tasks and have been successfully applied to image pair comparison. Daudt \emph{et al.}~\cite{daudt2018fully} were the first to introduce Fully Convolutional Networks (FCNs) into the binary change detection (BCD) field, developing several FCN architectures for better feature extraction. Following this, numerous CNN-based methods have been proposed~\cite{chen2019change, fang2021snunet, zhang2021escnet}. While these methods have achieved impressive results, the inherent limitations of the CNN architecture—specifically the restricted receptive field—hinder their ability to capture long-range dependencies, making them less effective when handling complex and diverse multi-temporal scenes with varying spatial-temporal resolutions.

Transformers, initially proposed by Vaswani \emph{et al.}~\cite{vaswani2017attention} for machine translation, have become the state-of-the-art method in many natural language processing tasks. Vision Transformers (ViTs)~\cite{dosovitskiy2020image} have shown significant success in visual representation learning, offering key advantages over CNNs by providing global context for each image patch through self-attention. This advantage has led to a surge of Transformer-based methods for change detection, such as the work by Chen \emph{et al.}~\cite{Chen2022Remote}, who were the first to apply Transformers to binary change detection (BCD). However, the quadratic complexity of self-attention in Transformers, especially with larger image sizes, increases computational costs and is problematic for dense prediction tasks like damage assessment and object detection in large-scale remote sensing datasets.
Similarly, in other domains like image generation, MCDM~\cite{shen2025long} proposes a motion-prior conditional diffusion model~\cite{shen2024boosting, shen2023advancing} for long-term TalkingFace generation, while their work on pose-guided person generation~\cite{shen2024imagpose} and virtual dressing~\cite{shen2024imagdressing} also highlights the challenge of balancing quality and efficiency in complex tasks. These advancements underline the broader challenge of handling computational complexity, which also impacts remote sensing change detection tasks.

In recent years, significant interest has grown in the State Space Model (SSM), originating from the Kalman filter model~\cite{Kalman1960}. The SSM concept was introduced in the S4 model~\cite{gu2021efficiently}, which can capture long-range dependencies and benefits from parallel training. Gu \emph{et al.}~\cite{gu2023mamba} proposed the Mamba architecture, which provides fast inference and linear scaling for sequence lengths, outperforming traditional models on real-world data with sequences up to millions of elements in length. Building on this, Zhu \emph{et al.}~\cite{zhu2024vision} introduced a new generic vision backbone called Vision Mamba (Vim). Recently, Mamba has been applied to the CD field, with Chen \emph{et al.}~\cite{chen2024changemamba} being the first to explore the potential of Mamba for remote sensing CD tasks. Zhang \emph{et al.}~\cite{zhang2024cdmamba} proposed CDMamba, a model that effectively integrates global and local features for better change detection. While Mamba, initially designed for 1D sequences, has been extended to vision tasks using various scanning patterns (e.g., spatially continuous or multiple simultaneous paths), these methods still rely on Mamba's core 1D scanning process. This 1D scanning approach leads to misrepresentations of geometric coherence in 2D images and spatial discrepancies (as illustrated in Fig.~\ref{fig:comparison}). Specifically, the way Mamba processes images as sequences fails to preserve the inherent 2D structure, resulting in distortions or loss of spatial relationships~\cite{zhang20242dmambaefficientstatespace}.

Feature Pyramid Network (FPN) is a deep learning framework used for object detection and image segmentation. FPN effectively utilizes information at different scales by constructing a pyramid structure of feature maps, thereby improving the accuracy of detection and segmentation. The FPN framework commonly relies on upsampling to enlarge smaller, semantically rich feature maps. However, bilinear upsampling, which interpolates uniformly sampled positions, only addresses fixed misalignments and is ineffective against more complex misalignments caused by residual connections and repeated downsampling/upsampling. Hence, dynamic position correspondence between feature maps is necessary to resolve these misalignments accurately.
To address the aforementioned challenges, we propose 2DMCG, a novel and efficient framework for remote sensing change detection. 2DMCG leverages the strengths of 2D Vision Mamba for robust feature extraction and introduces a change flow guidance mechanism derived from semantic flow to enhance the change decoding process.

\begin{figure*}[t]
    \centering
    \includegraphics[width=\linewidth]{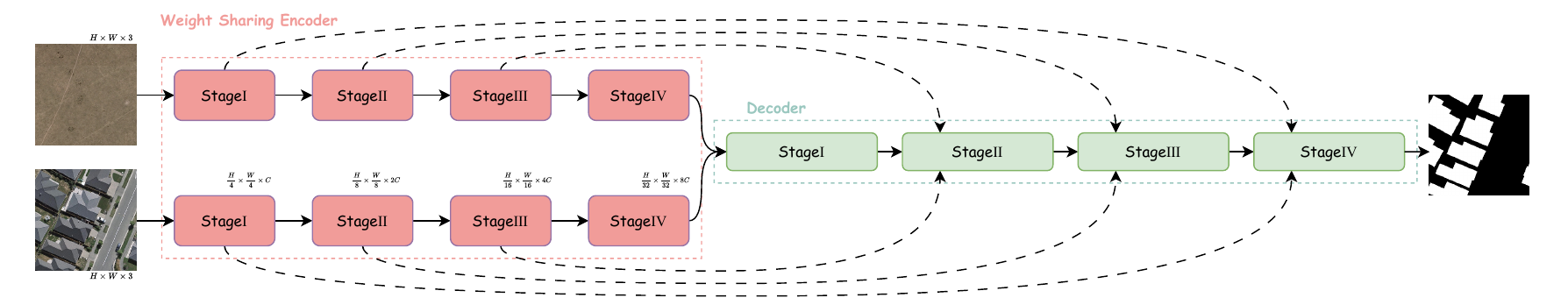}
    \caption{ \textbf{Illustration of the proposed change detection framework.} 
    The framework employs a Siamese architecture with shared weights for feature extraction. 
    Multi-temporal images are fed into the encoder to generate feature representations. 
    A change detection module then processes these features to produce the final change map. }
    \label{fig:overview}
\end{figure*}

Our contributions can be summarized as follows:
\begin{itemize}
    \item We introduce 2DMCG, a novel 2D Vision Mamba-based framework for remote sensing change detection, which overcomes the challenges of spatial misalignment and improves feature extraction.
    \item We incorporate a change flow guidance mechanism to enhance the decoding of change information, ensuring better spatial coherence and more accurate change detection.
    \item We demonstrate the superior performance of 2DMCG through extensive experiments on benchmark datasets, including LEVIR-CD+ and WHU-CD, where our framework significantly outperforms existing state-of-the-art methods.
\end{itemize}

\section{Related Work}\label{sec:rw}

\begin{figure*}[t] 
    \centering
    \begin{minipage}{0.98\textwidth} 
        \centering 
        \includegraphics[width=\textwidth]{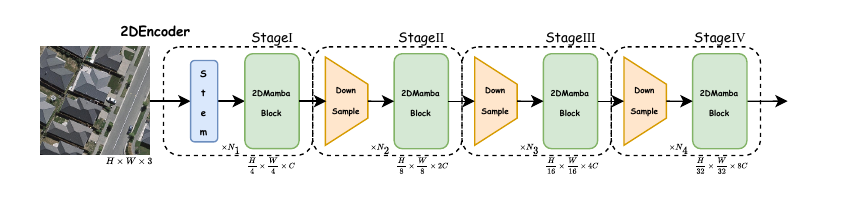} 
        \caption{ \textbf{Multi-stage encoder architecture based on 2D Mamba blocks.} 
        The encoder processes input images through multiple stages. 
        Each stage consists of a 2D Mamba block (repeated N times, where N1 to N4 are stage-specific repetition counts), 
        followed by a downsampling operation. 
        This design progressively reduces the spatial dimensions while extracting hierarchical features. }
    \label{fig:encoder}
    \end{minipage}
    \begin{minipage}{0.48\textwidth} 
        \centering 
        \includegraphics[width=0.6\textwidth, height=0.85\textwidth]{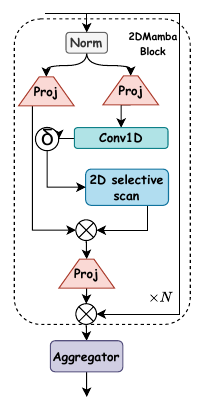}
    \subcaption{2D Mamba block structure}
    \label{fig:2dmamba}
    \end{minipage}
    \hfill 
    \begin{minipage}{0.48\textwidth} 
        \begin{algorithm}[H] 
        \small
            \caption{2D Selective Scan}
            \label{alg:2d_selective_scan}
            \begin{algorithmic}[1]
            \Require 2D input features $\mathbf{X} \in \mathbb{R}^{H \times W \times C}$; state dimension $N$; SSM parameters $\mathbf{A} \in \mathbb{R}^{N \times N}$, $\mathbf{B} \in \mathbb{R}^{1 \times N}$, and $\mathbf{C} \in \mathbb{R}^{N \times 1}$
            \Ensure 2D aggregated result $\mathbf{Y} \in \mathbb{R}^{H \times W}$
            \State Initialize $\mathbf{Y} \leftarrow \mathbf{0}$
            \For{$d \leftarrow 1$ to $N$}
                \State \Comment{Horizontal scan for state dimension $d$}
                \State $\mathbf{H}^{hor,d} \leftarrow \texttt{parallel\_horizontal\_scan}(\mathbf{A}, \mathbf{B}, \mathbf{X}, d)$
                \State \Comment{Vertical scan for state dimension $d$}
                \State $\mathbf{H}^d \leftarrow \texttt{parallel\_vertical\_scan}(\mathbf{A}, \mathbf{B}, \mathbf{H}^{hor,d}, d)$
                \State \Comment{Aggregate to the output}
                \State $\mathbf{Y} \leftarrow \mathbf{Y} + \mathbf{C} \mathbf{H}^d$
            \EndFor
            \State \Return $\mathbf{Y}$
            \end{algorithmic}
        \end{algorithm}
    \end{minipage}
    \caption{\textbf{Left}: The overall architecture of 2DMamba Block for feature representation. The 2D feature map is fed to $N$ layers of 2D-Mamba blocks.  \textbf{Right}: The 2D selective scan algorithm. It performs parallel horizontal scan and parallel vertical scan for each state dimension $d$ independently. Parameter $C$ then aggregates $N$ state dimensions into a single dimension output $y$.}
\end{figure*}

\subsection{CNN Based Method} 

\par In the early eras, Convolutional Neural Networks (CNNs)~\cite{lecun1998gradient} were regarded as the standard network design for computer vision tasks. As CNNs evolved, numerous architectures were proposed~\cite{he2016deep, huang2017densely, krizhevsky2012imagenet, simonyan2014very, szegedy2015going, xie2017aggregated} as vision backbones due to their excellent capability in extracting local features, which led to their widespread application in early change detection (CD) tasks. Daudt \emph{et al.}~\cite{daudt2018fully} first presented three Fully Convolutional Neural Network (FCNN) architectures that perform change detection on multi-temporal pairs of Earth observation images. Chen \emph{et al.}~\cite{chen2019change} proposed a novel and general deep Siamese Convolutional Multiple-Layers Recurrent Neural Network (SiamCRNN) for CD in multitemporal Very High Resolution (VHR) images. Fang \emph{et al.}~\cite{fang2021snunet} designed a densely connected Siamese network for change detection, namely SNUNet-CD, which combines a Siamese network and NestedUNet. This architecture refines and utilizes the most representative features of different semantic levels for the final classification. Zhang \emph{et al.}~\cite{zhang2021escnet} proposed an end-to-end superpixel-enhanced CD network (ESCNet) for VHR images, which combines differentiable superpixel segmentation and a deep convolutional neural network (DCNN).

\par Although these dominant follow-up works demonstrate superior performance and better efficiency, most of them still struggle to fully exploit long-distance dependencies~\cite{dosovitskiy2020image} due to the inherent local receptive field attributes of CNNs. This limitation is particularly significant in CD tasks, as change detection is optimized for spectral, spatial, temporal, and multi-sensor information representation. In this article, we introduce a new change detection method that overcomes the limitations of CNNs by explicitly modeling long-range dependencies, leading to improved representation of spectral, spatial and temporal information and consequently, more accurate change detection results.

\subsection{Transformer Based Method} 

\par The rapid evolution of Transformers~\cite{dosovitskiy2020image, shen2023git, liu2021Swin} in computer vision tasks has demonstrated immense potential for capturing long-range dependencies, significantly addressing the limitations faced by CNNs. Consequently, Transformer architectures~\cite{shen2023pbsl, shen2023triplet} have been introduced in the change detection (CD) field. Chen \emph{et al.}~\cite{Chen2022Remote} presented the first attempt to apply Transformers to binary change detection (BCD), efficiently and effectively modeling contexts within the spatial-temporal domain. Bandara \emph{et al.}~\cite{9883686} proposed a Transformer-based Siamese network architecture for change detection. Li \emph{et al.}~\cite{9761892} introduced an end-to-end encoding–decoding hybrid Transformer model for CD, combining the advantages of both Transformers and UNet. Additionally, Song \emph{et al.}~\cite{9961863} proposed a bi-branch fusion network based on axial cross attention to fuse local and global features.

\par However, despite the advantages of larger modeling capacity, which make the aforementioned Transformer-based methods perform well in CD, the number of visual tokens is limited due to the quadratic complexity of Transformers. This limitation leads to significant speed and memory costs when dealing with tasks involving long-range visual dependencies, such as CD. In this paper, we propose a novel 2D Vision Mamba-based framework for remote sensing change detection designed to overcome the computational complexity and memory footprint while maintaining model performance. 

\begin{figure}[t]
    \centering
    \includegraphics[width=\linewidth]{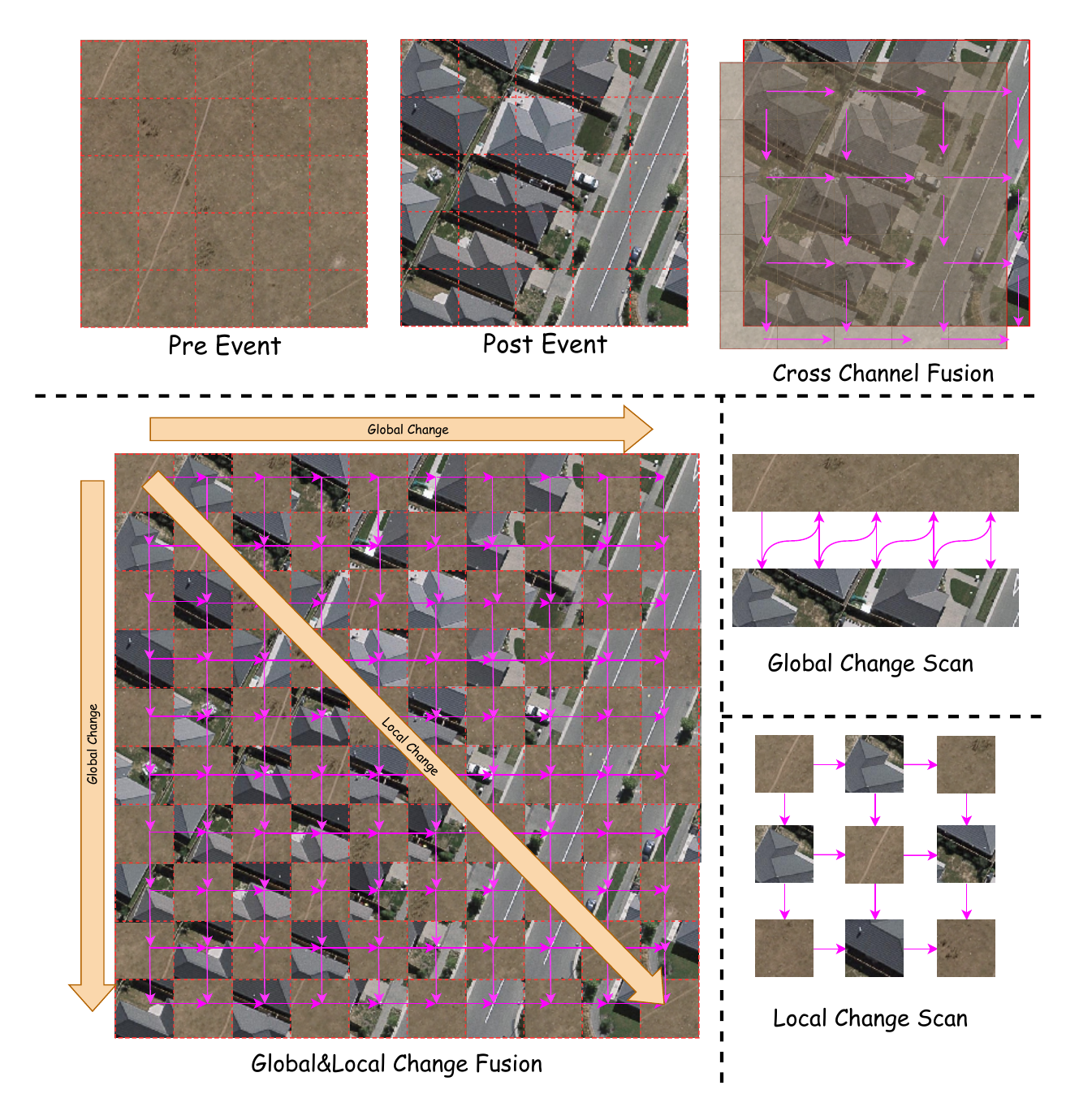}
    \caption{ \textbf{Illustration of two feature fusion methods for change detection.} 
    \textbf{Top:} \textit{Cross-Channel Fusion} concatenates pre- and post-event image features along the channel dimension and applies 2D Scan. 
    \textbf{Bottom:} \textit{Global \& Local Change Fusion} reorganizes the features into a larger map, enabling 2D Scan to capture both global changes (horizontal and vertical directions) and local changes (diagonal directions). }
    \label{fig:fusion}
\end{figure}

\subsection{State Space Based Method} 
\par The concept of the State Space Model (SSM) was first introduced in the S4 model~\cite{gu2021efficiently}, which demonstrated a promising ability to handle long-range dependencies both mathematically and empirically. Smith \emph{et al.}~\cite{smith2022simplified} introduced a new state space layer, the S5 layer, building on the design of the S4 layer. The S5 model revealed that a state space layer could leverage efficient and widely implemented parallel scans. Recently, based on the S4 model, Gu \emph{et al.}~\cite{gu2023mamba} proposed Mamba, which offers fast inference compared to Transformers, linear scaling in sequence length, and improved performance on real data up to million-length sequences. Mamba was soon introduced to computer vision tasks. Zhu \emph{et al.}~\cite{zhu2024vision} introduced a new generic vision backbone called Vision Mamba (Vim). This model marks image sequences with position embeddings and compresses the visual representation using bidirectional state space models, demonstrating significantly improved computation and memory efficiency. Ma \emph{et al.}~\cite{ma2024u} proposed U-Mamba, a general-purpose network for biomedical image segmentation inspired by State Space Sequence Models (SSMs).
\par In the field of change detection (CD), Chen \emph{et al.}~\cite{chen2024changemamba} explored for the first time the potential of the Mamba architecture for remote sensing CD tasks, fully utilizing its attributes to achieve spatio-temporal interaction of multi-temporal features, thereby obtaining accurate change information. Zhang \emph{et al.}~\cite{zhang2024cdmamba} proposed a model called CDMamba, which effectively combines global and local features for handling CD tasks.

\par However, the current formulations of these Mamba-based models are still limited to 1D and fail to fully utilize the 2D spatial information. In this paper, we apply a novel 2D Vision Mamba architecture which directly scans a 2D image without first flattening it into a 1D sequence. This is achieved through a hardware-aware 2D selective scan operator that extends the 1D Mamba parallelism into 2D, enabling efficient processing of spatial information.

\section{Proposed Method}\label{sec:method}
\subsection{Preliminaries}

\subsubsection{SSMs in Mamba and 1D Selective Scan}

State Space Models (SSMs) provide a function-to-function mapping for continuous systems, which, upon discretization, become sequence-to-sequence models. The discrete SSM dynamics are defined as:

\begin{align}
    \label{eq:discrete_ssm_state}
    \mathbf{h}_t &= \bar{\mathbf{A}}\mathbf{h}_{t-1} + \bar{\mathbf{B}}\mathbf{x}_t, \\
    \mathbf{y}_t &= \mathbf{C}\mathbf{h}_t = \sum_{d=1}^{N} \mathbf{C}^d \mathbf{h}_t^d.
    \label{eq:mamba_y}
\end{align}

Where $\mathbf{h}_t \in \mathbb{R}^N$ is the latent state vector at time $t$, $\mathbf{y}_t$ is the output vector, and $d \in \{1, 2, \dots, N\}$ indexes the state dimension. Traditional SSMs employ time-invariant matrices $\bar{\mathbf{A}}$ and $\bar{\mathbf{B}}$, which limits their ability to adapt to varying input contexts and effectively process long sequences.

To address this limitation, the Mamba block \cite{gu2023mamba} introduces a selective mechanism, enabling the SSM to dynamically adapt to the input context. This mechanism selectively aggregates relevant input information into the hidden state while discarding less important information. This selectivity is achieved by making the SSM parameters functions of the input:

\begin{equation}
    \label{eq:ssm}
    \begin{aligned}
        \bar{\mathbf{A}}_t &= \exp(\boldsymbol{\Delta}_t \mathbf{A}), &\quad \bar{\mathbf{B}}_t &= \boldsymbol{\Delta}_t \mathbf{B}(\mathbf{x}_t), \\
        \mathbf{C}_t &= \mathbf{C}(\mathbf{x}_t), &\quad \boldsymbol{\Delta}_t &= \mathrm{softplus}(\boldsymbol{\Delta}(\mathbf{x}_t)).
    \end{aligned}
\end{equation}

Where $\boldsymbol{\Delta}$, $\mathbf{B}$, and $\mathbf{C}$ are learnable linear functions of the input $\mathbf{x}_t$, and the diagonal matrix $\boldsymbol{\Delta}_t$ represents the discretized time step. This constitutes the 1D selective scan operation used in Mamba.

\subsubsection{2D Selective SSM Architecture}

Building upon the 1D selective scan, a 2D selective SSM architecture has been developed to process 2D feature maps directly, aggregating both geometric and semantic information. Unlike Mamba, which operates on flattened 1D sequences, this 2D approach employs parallel horizontal and vertical scans \cite{zhang20242dmambaefficientstatespace}. For simplicity, the state dimension superscript $d$ is omitted here. The parameterization of the 2D selective scan is consistent with the 1D case (Eq.~\eqref{eq:ssm}), with subscripts $(i, j)$ indexing 2D inputs instead of $t$. The input to the 2D selective scan, after normalization, projection, and convolution layers (as illustrated in Alg.~\ref{alg:2d_selective_scan}), is denoted as $x_{i,j}$.

The 2D selective scan comprises two steps:

\textbf{1. Horizontal Scan:} A 1D selective scan is applied independently to each row:

\begin{equation}
    \label{equ:h_scan}
    h_{i,j}^{\mathrm{hor}} = \bar{A}_{i, j} h_{i, j-1}^{\mathrm{hor}} + \bar{B}_{i, j} x_{i,j}.
\end{equation}

Where $h_{i, 0}^{\mathrm{hor}} = 0$, thus $h_{i,1}^{\mathrm{hor}} = \bar{B}_{i, 1} x_{i,1}$. The input-dependent parameters $\bar{A}_{i, j}$ and $\bar{B}_{i, j}$ modulate the influence of the previous horizontal state $h_{i, j-1}^{\mathrm{hor}}$ and the current input $x_{i,j}$.

\textbf{2. Vertical Scan:} Subsequently, a vertical scan is applied independently to each column of $h_{i,j}^{\mathrm{hor}}$. In this step, the term $\bar{B}_{i, j} x_{i,j}$ is replaced by the output of the horizontal scan, $h_{i,j}^{\mathrm{hor}}$:

\begin{equation}
    \label{equ:v_scan}
    h_{i,j} = \bar{A}_{i, j} h_{i-1, j} + h_{i,j}^{\mathrm{hor}}.
\end{equation}

With $h_{0, j} = 0$, resulting in $h_{1,j} = h^h_{1,j}$. The same parameter $\bar{A}_{i, j}$ is reused for the vertical scan.

Expanding Eqs.~\eqref{equ:h_scan} and \eqref{equ:v_scan} (and omitting the subscripts of $\bar{A}$ and $\bar{B}$ for notational simplicity), the hidden state $h_{i,j}$ can be expressed as the following recurrence:

\begin{equation}
    \label{equ:2d_recursion}
    h_{i,j} = \sum_{i'\leq i} \sum_{j' \leq j} \bar{A}^{(i - i' + j - j')} \bar{B}x_{i',j'}.
\end{equation}

Where $(i - i' + j - j')$ represents the Manhattan distance between $(i', j')$ and $(i, j)$, corresponding to a path from $(i', j')$ to $(i, j)$ traversing right horizontally and then down vertically. The final output $y_{i, j}$ is obtained by aggregating information from $h_{i, j}$ using a parameter $C$, analogous to 1D Mamba:

\begin{equation}
    y_{i, j} = C h_{i, j}.
\end{equation}

This 2D scanning mechanism aggregates information from all upper-left locations for each position $(i, j)$.

\subsubsection{Optical Flow}
Optical flow is widely used in video processing tasks \cite{zhu2017deep} to represent the apparent motion patterns of objects, surfaces, and edges in a visual scene caused by relative motion. Gadde \emph{et al.} \cite{gadde2017semantic} achieve video semantic segmentation by warping the internal features of the network. Nilsson \emph{et al.} \cite{nilsson2018semantic} warp the features of adjacent frames along the optical flow to predict the final segmentation map. Simonyan \emph{et al.} \cite{simonyan2014two} employ continuous multi-frame optical flow stacking for video action recognition. Furthermore, the concept of optical flow has also been incorporated into image semantic segmentation tasks. Li \emph{et al.} \cite{li2020semantic} propose the concept of semantic flow to align feature maps of different levels. In \cite{li2020improving}, the flow field is learned to warp image features and enhance the consistency of object features.

\subsection{Problem Statement}
\par This paper focuses on Binary Change Detection (BCD) within the Change Detection (CD) field. The task is defined as follows.

\par Binary Change Detection, a fundamental and extensively studied task in CD, identifies \textit{where} changes occur. BCD can be further categorized into category-agnostic CD, focusing on general land-cover changes, and single-category CD (e.g., building or forest CD). Given a training set $\mathcal{D}_{train}^{bcd} = \{(\mathbf{X}_i^{t_1}, \mathbf{X}_i^{t_2}, \mathbf{Y}_i^{bcd})\}_{i=1}^{N_{train}^{bcd}}$, where $\mathbf{X}_i^{t_1}, \mathbf{X}_i^{t_2} \in \mathbb{R}^{H \times W \times C}$ represent the $i$-th multi-temporal image pair acquired at times $t_1$ and $t_2$, respectively, and $\mathbf{Y}_i^{bcd} \in \{0, 1\}^{H \times W}$ is the corresponding binary change label, the objective of BCD is to train a change detector, $\mathcal{F}_{\theta}^{bcd}$, on $\mathcal{D}_{train}^{bcd}$ that accurately predicts binary change maps (change/no-change) for new image pairs.

\begin{figure*}[t]
    \centering
    \includegraphics[width=\linewidth]{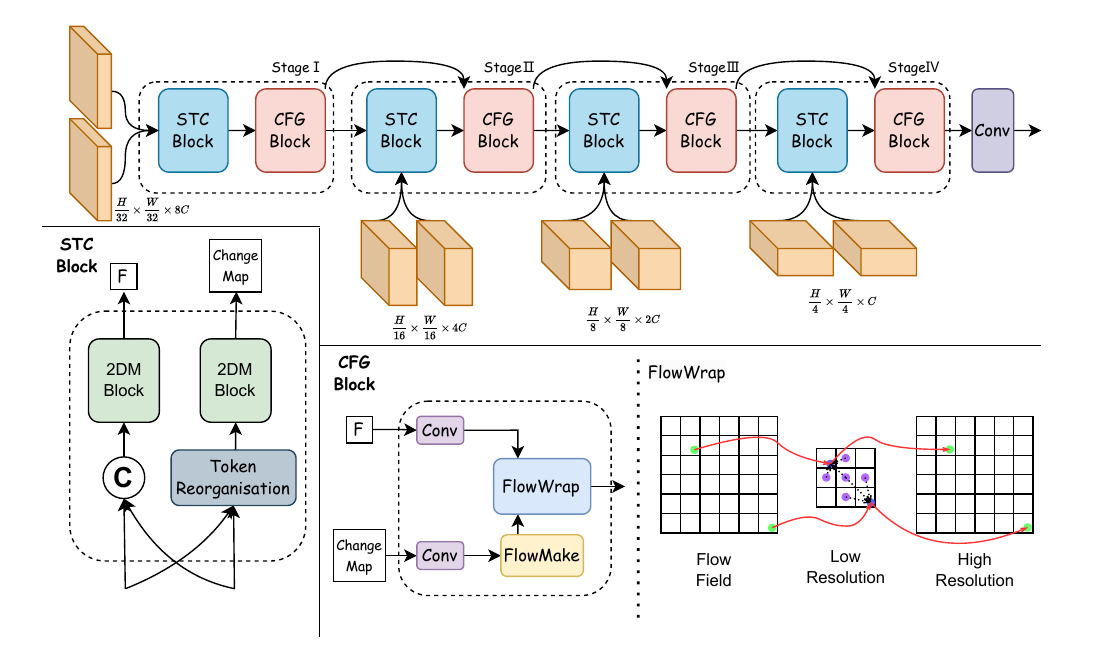}
    \caption{ \textbf{Overview of the proposed decoder architecture.} 
    \textbf{Top:} The complete decoder structure, showing the multi-stage design with STC (Spatial-Temporal Cross-Change) blocks and CFG (Change Flow Guided) blocks. 
    \textbf{Bottom:} Detailed illustration of the key modules within the decoder: STC block with 2DM (2D Mamba) blocks and Token Reorganization, CFG block with convolution and FlowWrap, and the FlowMake module align low-resolution change features to high-resolution ones. }
    \label{fig:decoder}
\end{figure*}

\subsection{Network Architecture}

This section details the architecture of the proposed network, illustrated in Fig. \ref{fig:overview}. The network is designed for Binary Change Detection (BCD) and comprises several key modules working in concert.

A Siamese encoder, denoted as $\mathcal{E}_{\theta}$, extracts multi-level features from bi-temporal input images $\mathbf{X}_i^{T_1}$ and $\mathbf{X}_i^{T_2}$. This process yields feature sets $\{\mathbf{E}_{i,j}^{T_1}\}_{j=1}^4 = \mathcal{E}_{\theta}(\mathbf{X}_i^{T_1})$ and $\{\mathbf{E}_{i,j}^{T_2}\}_{j=1}^4 = \mathcal{E}_{\theta}(\mathbf{X}_i^{T_2})$, where $j$ indexes the feature level (e.g., from shallow to deep). These feature sets are then input to a change decoder, $\mathcal{D}_{\theta}$, based on the 2DMamba architecture. The decoder effectively models spatio-temporal relationships between the bi-temporal features to generate a change probability map, $\mathbf{P}_i^{bcd} = \mathcal{D}_{\theta}(\{\mathbf{E}_{i,j}^{T_1}\}_{j=1}^4, \{\mathbf{E}_{i,j}^{T_2}\}_{j=1}^4)$. Finally, a binary change map, $\hat{\mathbf{Y}}_i^{bcd}$, is derived by selecting the class with the highest probability: $\hat{\mathbf{Y}}_i^{bcd} = \arg\max_c \mathbf{P}_i^{bcd}$.

The network architecture incorporates the following key modules:

\begin{itemize}
    \item \textbf{2D Encoder ($\mathcal{E}_{\theta}$):} This module employs a hierarchical architecture to extract features from input images with dimensions $H \times W \times 3$ (e.g., RGB). The encoder incorporates 2DMamba Blocks, specifically designed to capture spatially continuous features and effectively model spatial context.

    \item \textbf{Multi-Path 2D Cross-Fusion:} This module integrates features from different scales (levels) and potentially different paths within the encoder. A multi-path fusion strategy combines features from various encoder layers to capture both fine-grained and coarse-grained information. This fusion process enhances the representation of changes by considering information at multiple resolutions.

    \item \textbf{Change Flow Guided Decoder ($\mathcal{D}_{\theta}$):} This module generates the final change probability map $\mathbf{P}_i^{bcd}$ from the fused multi-level features. The decoder is designed to leverage change flow information (if explicitly computed or implicitly learned) to guide the decoding process and refine the localization of changed regions.
\end{itemize}

\subsection{2D Encoder}

The 2D encoder employs a hierarchical architecture for feature extraction from input images. As illustrated in Fig. \ref{fig:encoder}, it begins with an initial \textit{stem} module for preliminary feature extraction and channel adjustment. The encoder subsequently comprises four stages (Stages I-IV), each consisting of a 2D-Mamba Block followed by a Down Sample operation.

The core component of each stage is the 2D-Mamba Block, which processes 2D feature maps to capture spatial dependencies based on the Selective State Space Model (SSM) mechanism. The internal structure of the 2D-Mamba Block (detailed in Fig. \ref{fig:2dmamba}) involves the following steps: The input feature map is first normalized and then passed through two parallel linear projections. One projection is followed by a 1D convolution and a non-linear activation function $\delta$. The output of this pathway is then element-wise multiplied ($\otimes$) with the output of the other projection. This combined representation undergoes a further linear projection and is subsequently combined with the original normalized input through a residual connection using element-wise multiplication ($\otimes$). Finally, the outputs of $N$ consecutive 2D-Mamba blocks within each stage are aggregated by an Aggregator module.

Following each 2D-Mamba Block and Aggregator, a Down Sample operation is applied. This operation halves the spatial dimensions (both $H$ and $W$) while doubling the channel count. Consequently, the feature maps at each stage have increasing channel counts: $C$, $2C$, $4C$, and $8C$ for Stages I-IV, respectively. This hierarchical design enables the encoder to efficiently process 2D image data, capturing both local and long-range spatial dependencies while crucially maintaining the spatial continuity of the extracted features.

\subsection{Feature Fusion Modules}

This section describes two feature fusion modules designed to combine features from two time steps, $T_1$ and $T_2$: Cross-Channel Fusion (CCF) and Spatial Reorganization Fusion (SRF). These methods are illustrated in Figure \ref{fig:fusion}.

Let $\mathbf{F}_{i}^{T_1}$ and $\mathbf{F}_{i}^{T_2}$ represent the encoder output features at stage $i$ for time steps $T_1$ and $T_2$, respectively. Each feature map has dimensions $C \times H \times W$, where $C$ is the number of channels, $H$ is the height, and $W$ is the width.

\begin{itemize}
    \item \textbf{Cross-Channel Fusion (CCF):} This module concatenates the input features along the channel dimension, resulting in a feature map, $\mathbf{F}_{\text{ch}}$, with dimensions $2C \times H \times W$. The concatenation is defined as:

    \begin{equation}
        \mathbf{F}_{\text{ch}} = \text{Concat}(\mathbf{F}_{i}^{T_1}, \mathbf{F}_{i}^{T_2}).
    \end{equation}

    This method directly combines information from both time steps by increasing the channel dimension.

    \item \textbf{Spatial Reorganization Fusion (SRF):} 
    This module reorganizes the input features to create a new feature map, $\mathbf{F}_{2d}$, with dimensions $C \times 2H \times 2W$. The reorganization process can be described as follows:

    For the spatial dimensions of $\mathbf{F}_{2d}$ ($2H \times 2W$):

    \begin{align}
        \mathbf{F}_{2d}(2m, 2n) &= \mathbf{F}_{i}^{T_2}(m, n), \\
        \mathbf{F}_{2d}(2m+1, 2n) &= \mathbf{F}_{i}^{T_1}(m, n), \\
        \mathbf{F}_{2d}(2m, 2n+1) &= \mathbf{F}_{i}^{T_1}(m, n), \\
        \mathbf{F}_{2d}(2m+1, 2n+1) &= \mathbf{F}_{i}^{T_2}(m, n).
    \end{align}
    where $m \in \{0, 1, \ldots, H-1\}$ and $n \in \{0, 1, \ldots, W-1\}$.

    This reorganization can be compactly expressed as:

    \begin{equation}
        \mathbf{F}_{2d}(2m + a, 2n + b) =
        \begin{cases}
            \mathbf{F}_{i}^{T_1}(m, n) & \text{if } a \neq b, \\
            \mathbf{F}_{i}^{T_2}(m, n) & \text{if } a = b,
        \end{cases}
    \end{equation}
    where $a, b \in \{0, 1\}$.

    This reorganization method aims to capture both global and local changes. From horizontal and vertical perspectives, the reorganized feature $\mathbf{F}_{2d}$ represents bidirectional global changes. From the main diagonal and anti-diagonal perspectives, it represents bidirectional local changes.
\end{itemize}

\subsection{Change Flow Guided Decoder}

The proposed decoder (Fig.~\ref{fig:decoder}) comprises two core modules: the Spatial-Temporal Cross-Change (STC) module for bi-temporal feature fusion, and the Change Flow Guided Upsample (CFG) module for flow-guided upsampling.

\subsubsection{ChangeFlow: Learning Feature Correspondence}

Inspired by the success of optical flow in capturing motion \cite{zhu2017deep, gadde2017semantic, nilsson2018semantic, simonyan2014two}, ChangeFlow models the transformation of land surface features between time steps $T_1$ and $T_2$. Unlike optical flow, which describes motion in video frames, ChangeFlow focuses on feature-level correspondence in bi-temporal remote sensing imagery.  Given pre-change image $I_1$ and post-change image $I_2$ (Fig.~\ref{fig:flow}, columns 1-2), ChangeFlow estimates a flow field $\mathcal{F}$ mapping features from $I_1$ to $I_2$.  This facilitates direct feature comparison in a shared space for accurate change detection.  Analogous to semantic flow \cite{li2020semantic, li2020improving}, ChangeFlow enhances feature consistency across time.  Specifically, $\mathcal{F}$ warps features of $I_1$ towards $I_2$, enabling the network to identify changed regions (Fig.~\ref{fig:flow}, last column). The ground truth change mask is shown in Fig.~\ref{fig:flow}, third column.  Thermal activation maps (Fig.~\ref{fig:flow}, fourth column) highlight the network's focus on change-related features.

\subsubsection{Change Flow Guided Upsample (CFG)}

The CFG module employs a Feature Pyramid Network (FPN)-like structure. Feature maps at each level are channel-compressed using two $1 \times 1$ convolutions before being passed to the next level.  Given feature maps $\mathbf{F}_{2di}$ ($2H \times 2W$) from the STC module and $\mathbf{F}_{chi}$ ($H \times W$) from the SRF module, $\mathbf{F}_{2di}$ is processed by two $3 \times 3$ convolutional layers to predict a change flow field $\Delta_{i-1} \in \mathbb{R}^{H \times W \times 2}$.

The flow field $\Delta_{i-1}$ maps each position $p_{i-1}$ on the spatial grid $\Omega_{i-1}$ to a corresponding point $p_i$ at the next higher resolution level $i$:

\begin{equation}
    p_{i} = \frac{p_{i-1}+\Delta_{i-1}(p_{i-1})}{2}.
    \label{eq:mapping}
\end{equation}

This mapping accounts for the resolution difference (Fig.~\ref{fig:decoder}).

Warped features $\widetilde{\mathbf{F}}_i$ at locations $p_{i-1}$ are then obtained via bilinear interpolation \cite{STN}:

\begin{equation}
    \widetilde{\mathbf{F}}_i(p_{i-1}) = \sum_{p \in \mathcal{N}(p_{i})} w_p \mathbf{F}_{i}(p).
    \label{eq:interpolate}
\end{equation}

Where $\mathcal{N}(p_{i})$ are the four neighbors of $p_i$ in $\mathbf{F}_i$, and $w_p$ are the bilinear interpolation weights.

\begin{figure}[t]
    \centering
    \begin{minipage}[b]{0.45\textwidth}
        \centering
        \includegraphics[width=\textwidth]{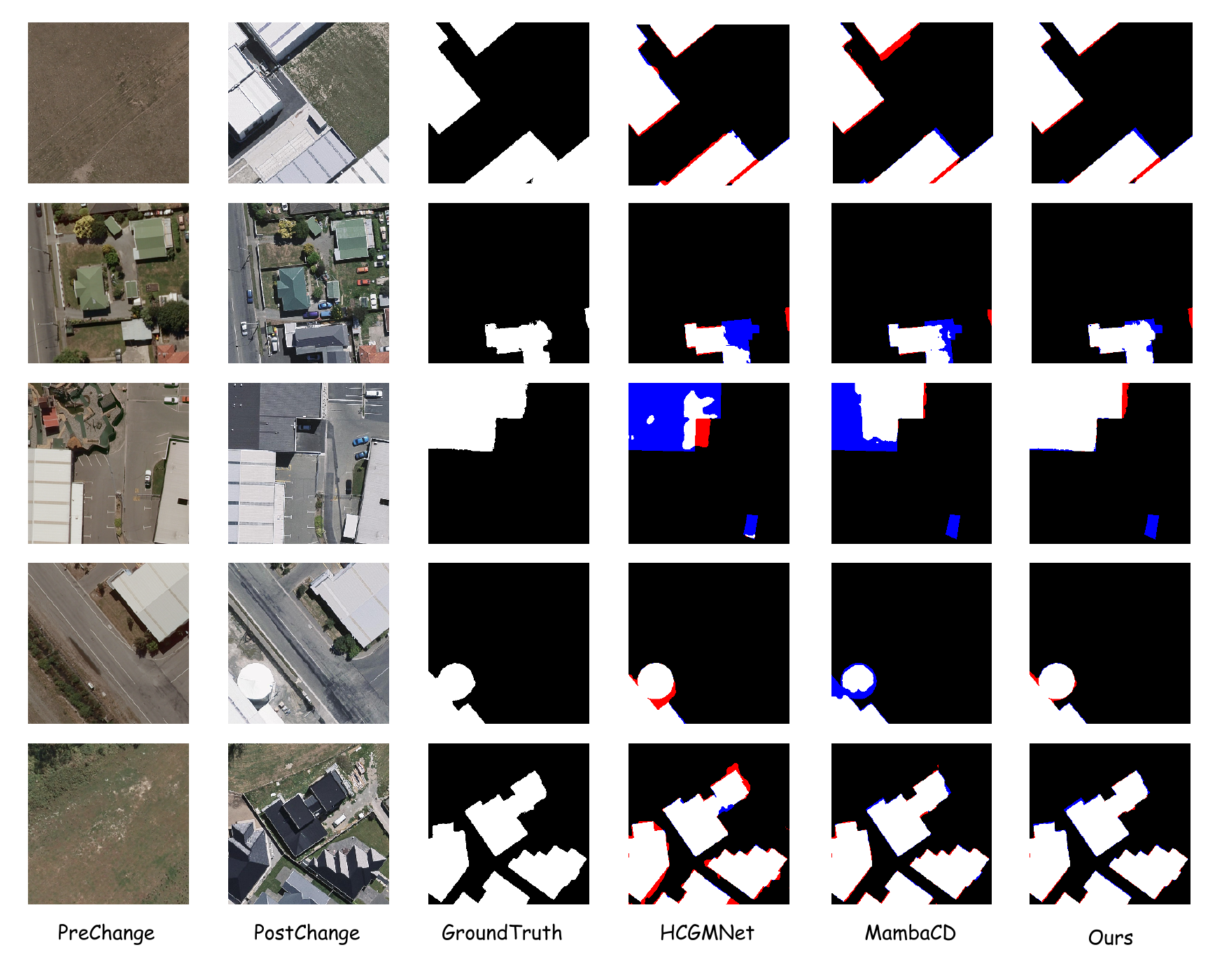}
        \subcaption{Visualization results of different change detection methods on the WHU-CD test set. In the visualizations, white represents true positives, black represents true negatives, \textcolor{red}{red} indicates false positives, and \textcolor{blue}{blue} indicates false negatives.}
        \label{fig:whu}
    \end{minipage}
    \hfill
    \begin{minipage}[b]{0.45\textwidth}
        \centering
        \includegraphics[width=\textwidth]{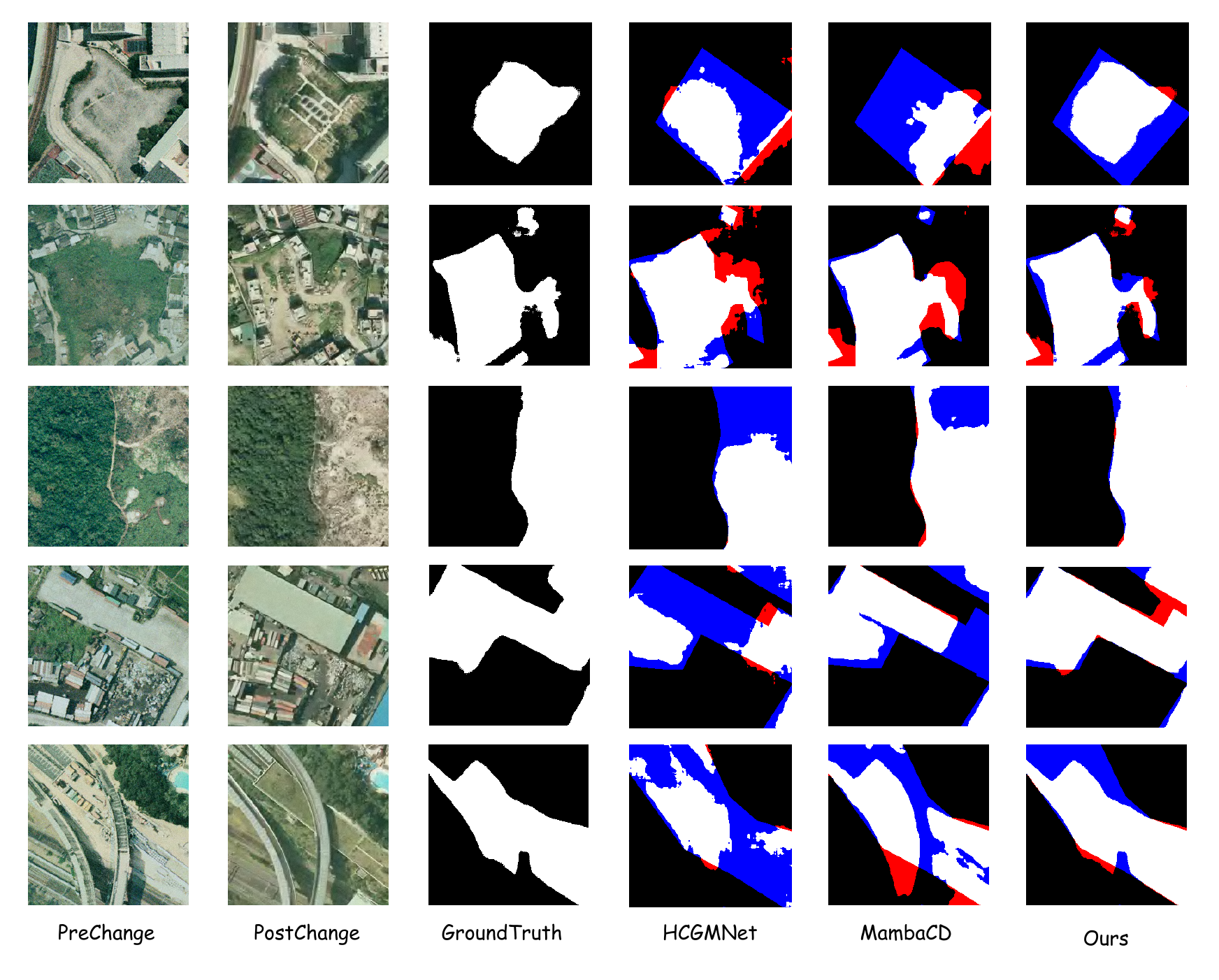}
        \subcaption{Visualization results of different change detection methods on the SYSU-CD test set. In the visualizations, white represents true positives, black represents true negatives, \textcolor{red}{red} indicates false positives, and \textcolor{blue}{blue} indicates false negatives.}
        \label{fig:sysu}
    \end{minipage}
    \caption{Comparison of change detection results on two different datasets. (a) shows the results on the WHU-CD test set, while (b) presents the results on the SYSU-CD test set. In both visualizations, white represents true positives, black represents true negatives, red indicates false positives, and blue indicates false negatives. These visualizations help in understanding the performance of different change detection methods on diverse datasets.}
    \label{fig:whu_and_sysu}
\end{figure}

\section{Experiment and Analysis}\label{sec:exp} 

To validate the proposed 2DMCG method's superiority, it is compared with multiple state-of-the-art approaches on three large-scale datasets, namely, WHU-CD, SYSU and LEVIR-CD+.

\subsection{Datasets}
\textbf{\emph{WHU-CD}~\cite{ji2018fully}}
The WHU-CD dataset, a subset of the larger WHU Building dataset, is specifically tailored for building change detection (CD) tasks. It consists of a pair of high-resolution spatial remote sensing images of Christchurch, New Zealand, captured in April 2012 and April 2016. The images have a spatial resolution of 0.2 meters/pixel and cover an area of 20.5 square kilometers. The 2012 dataset features 12,796 buildings, while the 2016 dataset shows an increase to 16,077 buildings within the same area, reflecting significant urban development over the four-year period. This dataset is particularly focused on detecting changes in large and sparse building structures.

\textbf{\emph{SYSU-CD}~\cite{shi2021deeply}}
This dataset is a category-agnostic change detection (CD) dataset, comprising a comprehensive collection of 20,000 pairs of aerial images with a resolution of 0.5 meters per pixel, captured in Hong Kong between 2007 and 2014. It is notable for its emphasis on urban and coastal transformations, including high-rise buildings and infrastructure developments. The dataset covers a wide array of change scenarios, such as urban construction, suburban expansion, groundwork, vegetation changes, road expansion, and sea construction.

\textbf{\emph{LEVIR-CD+}~\cite{Chen2020}}
The LEVIR-CD+ dataset is an enhanced version of the LEVIR-CD, specifically designed for urban building change detection using RGB image pairs sourced from Google Earth. It comprises 985 image pairs, each with dimensions of 1024 $\times$ 1024 pixels and a spatial resolution of 0.5 meters per pixel. This dataset includes masks for building and land use changes across 20 different regions in Texas, covering the period from 2002 to 2020, with observations taken at 5-year intervals. LEVIR-CD+ is considered a more accessible version of the S2Looking dataset, largely due to its focus on urban areas and near-nadir viewing angles.

\begin{figure}[t]
    \centering
\includegraphics[width=0.95\linewidth]{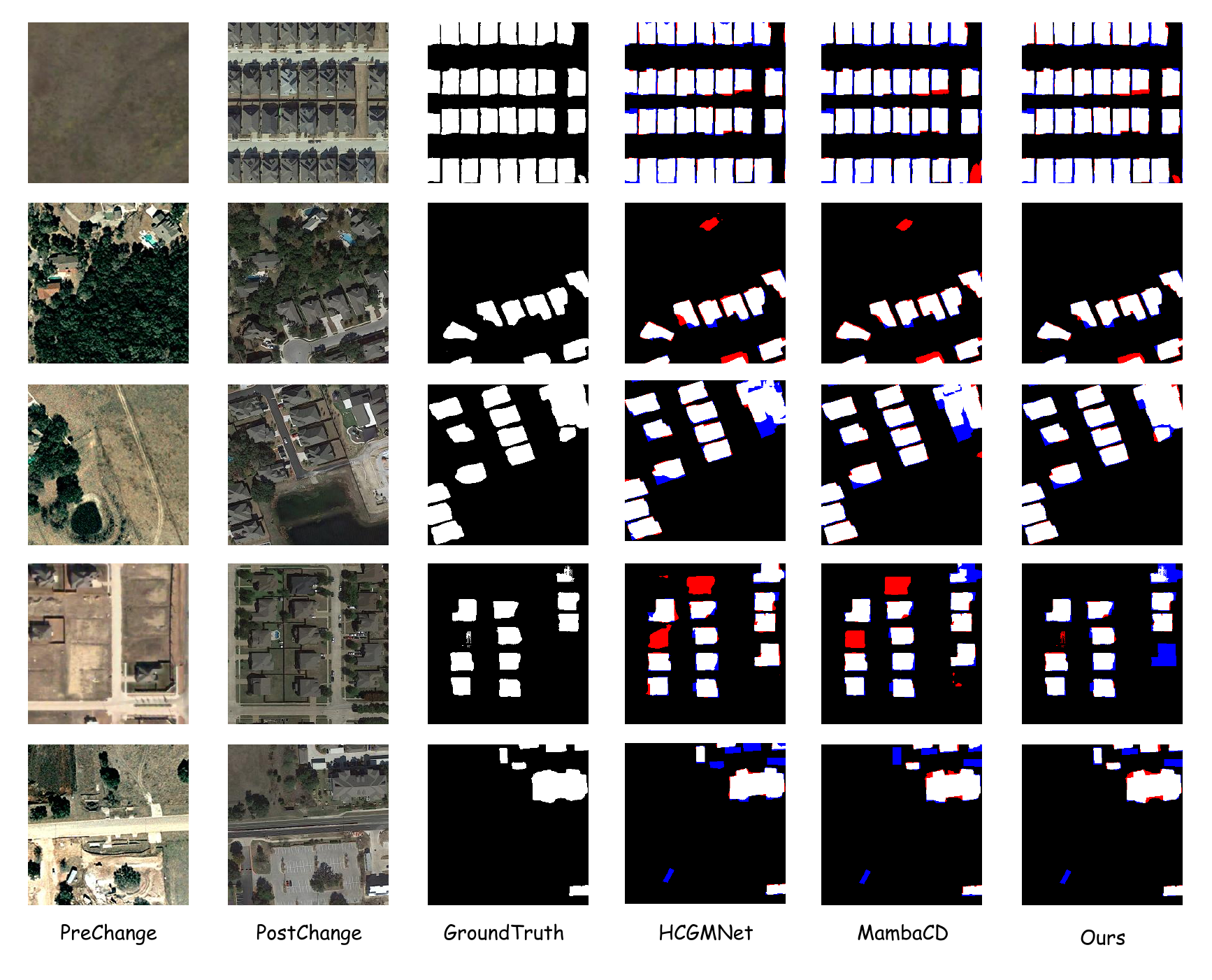}
    \caption{Visualization results of different change detection methods on the LEVIR-CD+ test set. In the visualizations, white represents true positives, black represents true negatives, \textcolor{red}{red} indicates false positives, and \textcolor{blue}{blue} indicates false negatives.}
    \label{fig:levir+}
\end{figure}

\subsection{Evaluation Metrics} 
\par To assess the effectiveness of the proposed 2DMCG, we utilized five primary evaluation metrics: overall accuracy (OA), precision (Pre), recall (Rec), F1 score, and intersection over union (IoU). Overall accuracy (OA) indicates the ratio of correctly predicted pixels to the total number of pixels. Precision (Pre) measures the proportion of true positive pixels among all pixels identified as positive. Recall (Rec) quantifies the proportion of true positive pixels relative to all actual positive pixels in the ground truth. The F1 score provides a balance between precision and recall by computing their harmonic mean. Intersection over union (IoU) evaluates the overlap between the predicted positive regions and the ground truth. These metrics are defined as follows.

\begin{align}
    OA = \frac{TP+TN}{TP+TN+FP+FN}, Precision = \frac{TP}{TP+FP}, Recall = \frac{TP}{TP+FN},\\ F1 = \frac{2}{Recall^{-1}+Precision^{-1}}, IoU = \frac{TP}{TP+FP+FN}.
\end{align}

Where TP, TN, FP, and FN denote the counts of true positives, true negatives, false positives, and false negatives, respectively.

\subsection{Implementation Details} 
\par The proposed 2DMCG is implemented using the Pytorch framework and executed on an NVIDIA A100 GPU. For optimization, we employ the Adam optimizer with an initial learning rate of 1$e$-4. The parameters ${\beta_1}$ and ${\beta_2}$ are set to 0.9 and 0.999, respectively. The batch size is configured to 8, and the total number of training step is 30000. The loss function is a combination of cross-entropy loss and dice loss.

\subsection{Comparison with State-of-the-art Methods}

This section evaluates the performance of our proposed method ("Ours") against a diverse set of state-of-the-art change detection techniques across three benchmark datasets: WHU-CD, SYSU-CD, and LEVIR-CD+.  The comparison includes representative methods from CNN-based (FC-EF~\cite{daudt2018fully}, FC-Siam-Diff~\cite{daudt2018fully}, FC-Siam-Conc~\cite{daudt2018fully}, SNUNet~\cite{fang2021snunet}, HANet~\cite{han2023hanet}, CGNet~\cite{Han2023CGNet}, SEIFNet~\cite{Huang2024Spatiotemporal}), Transformer-based (ChangeFormer~\cite{9883686}, BIT~\cite{Chen2022Remote}, TransUNetCD~\cite{li2022transunetcd}, SwinSUNet~\cite{zhang2022swinsunet}, CTDFormer~\cite{zhang2023relation}), and Mamba-based (ChangeMamba~\cite{chen2024changemamba}) architectures. Performance is assessed using Recall (Rec), Precision (Prec), Overall Accuracy (OA), F1-score (F1), Intersection over Union (IoU), and Kappa Coefficient (KC).  The results, with the top two performers in each metric highlighted in \textcolor{red}{red} (best) and \textcolor{blue}{blue} (second best), are detailed below for each dataset.

\subsubsection{Comparisons on WHU-CD}

\begin{table}[t]
\centering
\caption{Accuracy assessment for different binary CD models on the WHU-CD dataset.}\label{tab:results-whu}
\scriptsize
\resizebox{0.85\linewidth}{!}{
\begin{tabular}{l|r|r|r|r|r|r}\toprule
Model &Rec &Precision &OA &F1 &IoU &KC \\\cmidrule{1-7}
FC-EF~\cite{daudt2018fully} &86.33 &83.50 &98.87 &84.89 &73.74 &84.30 \\\cmidrule{1-7}
FC-Siam-Diff~\cite{daudt2018fully} &84.69 &90.86 &99.13 &87.67 &78.04 &87.22 \\\cmidrule{1-7}
FC-Siam-Conc~\cite{daudt2018fully} &87.72 &84.02 &98.94 &85.83 &75.18 &85.28 \\\cmidrule{1-7}
SiamCRNN-18~\cite{chen2019change} &90.48 &91.56 &99.34 &91.02 &83.51 &90.68 \\\cmidrule{1-7}
SiamCRNN-101~\cite{chen2019change} &90.45 &87.79 &99.19 &89.10 &80.34 &88.68 \\\cmidrule{1-7}
SNUNet~\cite{fang2021snunet} &87.36 &88.04 &99.10 &87.70 &78.09 &87.23 \\\cmidrule{1-7}
DSIFN~\cite{zhang2020deeply} &83.45 &\textcolor{red}{97.46} &99.31 &89.91 &81.67 &89.56 \\\cmidrule{1-7}
HANet~\cite{han2023hanet} &88.30 &88.01 &99.16 &88.16 &78.82 &87.72 \\\cmidrule{1-7}
CGNet~\cite{Han2023CGNet} &90.79 &94.47 &99.48 &92.59 &86.21 &92.33 \\\cmidrule{1-7}
SEIFNet~\cite{Huang2024Spatiotemporal} &90.66 &91.93 &99.36 &91.29 &83.98 &90.96 \\\cmidrule{1-7}
ChangeFormerV1~\cite{9883686} &84.30 &90.80 &99.11 &87.43 &77.67 &86.97 \\\cmidrule{1-7}
ChangeFormerV6~\cite{9883686} &81.90 &85.49 &98.83 &83.66 &71.91 &83.05 \\\cmidrule{1-7}
BIT-18~\cite{Chen2022Remote} &90.36 &90.30 &99.29 &90.33 &82.37 &89.96 \\\cmidrule{1-7}
BIT-101~\cite{Chen2022Remote} &90.24 &89.83 &99.27 &90.04 &81.88 &89.66 \\\cmidrule{1-7}
TransUNetCD~\cite{li2022transunetcd} &90.50 &85.48 &99.09 &87.79 &78.44 &87.44 \\\cmidrule{1-7}
SwinSUNet~\cite{zhang2022swinsunet} &92.03 &94.08 &99.50 &93.04 &87.00 &92.78 \\\cmidrule{1-7}
CTDFormer~\cite{zhang2023relation} &85.37 &92.23 &99.20 &88.67 &79.65 &88.26 \\\cmidrule{1-7}
MambaBCD-Base~\cite{chen2024changemamba} &\textcolor{blue}{92.24} &96.16 &\textcolor{red}{99.56} &\textcolor{blue}{94.20} &\textcolor{blue}{89.01} &\textcolor{blue}{93.92} \\\cmidrule{1-7}
Ours &\textcolor{red}{93.69} &\textcolor{blue}{96.48} &\textcolor{blue}{99.53} &\textcolor{red}{95.07} &\textcolor{red}{90.59} &\textcolor{red}{94.81} \\\cmidrule{1-7}
\bottomrule
\end{tabular}
}
\end{table}

Table \ref{tab:results-whu} presents the results on the WHU-CD dataset.  Our method achieves top performance in Recall (93.69\%), F1-score (95.07\%), IoU (90.59\%), and KC (94.81\%), demonstrating its effectiveness in accurately delineating changed regions with minimal false positives.  While MambaBCD-Base achieves the highest OA (99.56\%), our method's superior performance in other critical metrics indicates a better balance between detection accuracy and precision.  Although other deep learning methods (SiamCRNN, SNUNet, DSIFN, HANet, CGNet, SEIFNet, ChangeFormer, BIT, TransUNetCD, SwinSUNet, CTDFormer) achieve competitive results, particularly in OA, our approach exhibits a clear overall advantage.

\subsubsection{Comparisons on SYSU-CD}

\begin{table}[t]\centering
\caption{Accuracy assessment for different binary CD models on the SYSU-CD dataset.}\label{tab:results-sysu}
\scriptsize
\resizebox{.85\linewidth}{!}{
\begin{tabular}{l|r|r|r|r|r|r}\toprule
Model &Rec &Precision &OA &F1 &IoU &KC \\\cmidrule{1-7}
FC-EF~\cite{daudt2018fully} &75.17 &76.47 &88.69 &75.81 &61.04 &68.43 \\\cmidrule{1-7}
FC-Siam-Diff~\cite{daudt2018fully} &75.30 &76.28 &88.65 &75.79 &61.01 &68.38 \\\cmidrule{1-7}
FC-Siam-Conc~\cite{daudt2018fully} &76.75 &73.67 &88.05 &75.18 &60.23 &67.32 \\\cmidrule{1-7}
SiamCRNN-18~\cite{chen2019change} &76.83 &84.80 &91.29 &80.62 &67.54 &75.02 \\\cmidrule{1-7}
SiamCRNN-101~\cite{chen2019change} &\textcolor{blue}{80.48} &80.40 &90.77 &80.44 &67.28 &74.40 \\\cmidrule{1-7}
SNUNet~\cite{fang2021snunet} &72.21 &74.09 &87.49 &73.14 &57.66 &64.99 \\\cmidrule{1-7}
DSIFN~\cite{zhang2020deeply} &\textcolor{red}{82.02} &75.83 &89.59 &78.80 &65.02 &71.92 \\\cmidrule{1-7}
HANet~\cite{han2023hanet} &76.14 &78.71 &89.52 &77.41 &63.14 &70.59 \\\cmidrule{1-7}
CGNet~\cite{Han2023CGNet} &74.37 &\textcolor{red}{86.37} &91.19 &79.92 &66.55 &74.31 \\\cmidrule{1-7}
SEIFNet~\cite{Huang2024Spatiotemporal} &78.29 &78.61 &89.86 &78.45 &64.54 &71.82 \\\cmidrule{1-7}
ChangeFormerV1~\cite{9883686} &75.82 &79.65 &89.73 &77.69 &63.52 &71.02 \\\cmidrule{1-7}
ChangeFormerV6~\cite{9883686} &72.38 &81.70 &89.67 &76.76 &62.29 &70.15 \\\cmidrule{1-7}
BIT-18~\cite{Chen2022Remote} &76.42 &\textcolor{blue}{84.85} &91.22 &80.41 &67.24 &74.78 \\\cmidrule{1-7}
BIT-101~\cite{Chen2022Remote} &75.58 &83.64 &90.76 &79.41 &65.84 &73.47 \\\cmidrule{1-7}
TransUNetCD~\cite{li2022transunetcd} &77.73 &82.59 &90.88 &80.09 &66.79 &74.18 \\\cmidrule{1-7}
SwinSUNet~\cite{zhang2022swinsunet} &79.75 &83.50 &91.51 &\textcolor{red}{81.58} &\textcolor{red}{68.89} &76.06 \\\cmidrule{1-7}
CTDFormer~\cite{zhang2023relation} &75.53 &80.80 &90.00 &78.08 &64.04 &71.61 \\\cmidrule{1-7}
MambaBCD-Base~\cite{chen2024changemamba} &80.01 &82.61 &\textcolor{blue}{92.07} &\textcolor{blue}{81.29} &\textcolor{blue}{68.47} &\textcolor{blue}{76.26} \\\cmidrule{1-7}
Ours &78.09 &84.64 &\textcolor{red}{92.24} &81.23 &68.40 &\textcolor{red}{76.34} \\\cmidrule{1-7}
\bottomrule
\end{tabular}
}
\end{table}

Table \ref{tab:results-sysu} summarizes the results on the SYSU-CD dataset.  Similar to the WHU-CD results, both our method and MambaBCD-Base demonstrate strong performance. Our method attains the highest OA (92.24\%) and KC (76.34\%), along with competitive F1-score (81.23\%) and IoU (68.40\%). MambaBCD-Base achieves the highest Recall (82.02\%) and competitive results across other metrics, highlighting its ability to capture a large portion of actual changes.  SwinSUNet also performs well, particularly in F1-score (81.58\%) and IoU (68.89\%).  The performance variations across metrics underscore the importance of considering multiple evaluation criteria on this dataset, which presents a significant challenge for change detection.

\subsubsection{Comparisons on LEVIR-CD+}

\begin{table}[t]\centering
\caption{Accuracy assessment for different binary CD models on the LEVIR-CD+ dataset.}\label{tab:levir+}
\scriptsize
\resizebox{.85\linewidth}{!}{
\begin{tabular}{l|r|r|r|r|r|r}\toprule
Model &Rec &Precision &OA &F1 &IoU &KC \\\cmidrule{1-7}
FC-EF~\cite{daudt2018fully} &71.77 &69.12 &97.54 &70.42 &54.34 &69.14 \\\cmidrule{1-7}
FC-Siam-Diff~\cite{daudt2018fully} &74.02 &81.49 &98.26 &77.57 &63.36 &76.67 \\\cmidrule{1-7}
FC-Siam-Conc~\cite{daudt2018fully} &78.49 &78.39 &98.24 &78.44 &64.53 &77.52 \\\cmidrule{1-7}
SiamCRNN-18~\cite{chen2019change} &84.25 &81.22 &98.56 &82.71 &70.52 &81.96 \\\cmidrule{1-7}
SiamCRNN-101~\cite{chen2019change} &80.96 &85.56 &98.67 &83.20 &71.23 &82.50 \\\cmidrule{1-7}
SNUNet~\cite{fang2021snunet} &78.73 &71.07 &97.83 &74.70 &59.62 &73.57 \\\cmidrule{1-7}
DSIFN~\cite{zhang2020deeply} &84.36 &83.78 &98.70 &84.07 &72.52 &83.39 \\\cmidrule{1-7}
HANet~\cite{han2023hanet} &75.53 &79.70 &98.22 &77.56 &63.34 &76.63 \\\cmidrule{1-7}
CGNet~\cite{Han2023CGNet} &\textcolor{blue}{86.02} &81.46 &98.63 &83.68 &71.94 &82.97 \\\cmidrule{1-7}
SEIFNet~\cite{Huang2024Spatiotemporal} &81.86 &84.83 &98.66 &83.32 &71.41 &83.63 \\\cmidrule{1-7}
ChangeFormerV1~\cite{9883686} &77.00 &82.18 &98.38 &79.51 &65.98 &78.66 \\\cmidrule{1-7}
ChangeFormerV6~\cite{9883686} &78.57 &67.66 &97.60 &72.71 &57.12 &71.46 \\\cmidrule{1-7}
BIT-18~\cite{Chen2022Remote} &80.86 &83.76 &98.58 &82.28 &69.90 &81.54 \\\cmidrule{1-7}
BIT-101~\cite{Chen2022Remote} &81.20 &83.90 &98.60 &82.53 &70.26 &81.80 \\\cmidrule{1-7}
TransUNetCD~\cite{li2022transunetcd} &84.18 &83.08 &98.66 &83.63 &71.86 &82.93 \\\cmidrule{1-7}
SwinSUNet~\cite{zhang2022swinsunet} &85.85 &85.34 &98.92 &85.60 &74.82 &84.98 \\\cmidrule{1-7}
CTDFormer~\cite{zhang2023relation} &80.03 &80.58 &98.40 &80.30 &67.09 &79.47 \\\cmidrule{1-7}
MambaBCD-Base~\cite{chen2024changemamba} &\textcolor{red}{86.43} &\textcolor{blue}{88.80} &\textcolor{blue}{99.00} &\textcolor{blue}{87.60} &\textcolor{blue}{77.94} &\textcolor{blue}{87.08} \\\midrule
Ours &85.25 &\textcolor{red}{90.41} &\textcolor{red}{99.04} &\textcolor{red}{87.75} &\textcolor{red}{78.18} &\textcolor{red}{87.25} \\
\bottomrule
\end{tabular}
}
\end{table}

Table \ref{tab:levir+} presents the results on the LEVIR-CD+ dataset.  Our proposed method ("Ours") again demonstrates excellent performance, achieving the highest scores in Precision (90.41\%), OA (99.04\%), F1-score (87.75\%), IoU (78.18\%), and KC (87.25\%). These results highlight its ability to accurately identify change regions while minimizing both false positives and false negatives. While MambaBCD-Base exhibits competitive performance, particularly in Recall (86.43\%), our method's superior Precision translates to better F1-score and IoU. This suggests that our approach is more effective at distinguishing actual changes from spurious ones, a critical factor in real-world applications.  The table also reflects the general trend of deep learning-based methods outperforming traditional approaches.  Siamese architectures, convolutional recurrent networks, and Transformer-based models all achieve competitive results, showcasing the advancements in deep learning for change detection.  Overall, the results on LEVIR-CD+ further validate the effectiveness of our proposed method.

\subsection{Ablation Studies and Analysis}  
This section presents ablation studies conducted to analyze the contribution of the key components of our proposed method.  Specifically, we investigate the impact of the change flow guidance mechanism and the 2D Mamba Scan (2DS) on the overall performance.  The experiments are conducted across three benchmark datasets: WHU-CD, SYSU-CD, and LEVIR-CD+.  Performance is evaluated using Recall (Rec), Precision (Prec), Overall Accuracy (OA), F1-score (F1), Intersection over Union (IoU), and Kappa Coefficient (KC). The ablation experiments follow a consistent setup: we compare the full proposed method against two ablated versions:

\begin{table}[t]
\centering
\caption{Ablation Study Results on WHU-CD, SYSU-CD, and LEVIR-CD+ datasets.}
\label{tab:ab}
\scriptsize
\resizebox{.85\linewidth}{!}{
\begin{tabular}{l|c|c|c|c|c|c|c|c}
\toprule
Dataset  & ChangeFlow & 2DS & Rec & Precision & OA & F1 & IoU & KC \\
\midrule
\multirow{3}{*}{WHU}  & \checkmark & \checkmark & 93.69 & 96.48 & 99.53 & 95.07 & 90.59 & 94.81 \\
 & $\times$ & \checkmark & 89.96 & 94.55 & 99.25 & 92.20 & 85.53 & 91.80 \\
 & \checkmark & $\times$ & 91.48 & 96.49 & 99.41 & 93.92 & 88.54 & 93.61 \\
\midrule
\multirow{3}{*}{SYSU}  & \checkmark & \checkmark & 78.09 & 84.64 & 92.24 & 81.23 & 68.40 & 76.34 \\
 & $\times$ & \checkmark & 77.39 & 78.62 & 90.60 & 78.00 & 63.93 & 72.02 \\
 & \checkmark & $\times$ & 79.93 & 81.41 & 91.75 & 80.66 & 67.59 & 75.42 \\
\midrule
\multirow{3}{*}{LEVIR+}  & \checkmark & \checkmark & 85.25 & 90.41 & 99.04 & 87.75 & 78.18 & 87.25 \\
 & $\times$ & \checkmark & 83.10 & 83.17 & 98.62 & 83.13 & 71.14 & 82.42 \\
 & \checkmark & $\times$ & 84.70 & 88.65 & 98.93 & 86.63 & 76.41 & 86.07 \\
\bottomrule
\end{tabular}
}
\end{table}

\begin{itemize}
    \item \textbf{w/o Flow}: This version removes the change flow guidance during feature fusion and the decoding process. This ablation aims to evaluate the effectiveness of incorporating change flow information to guide feature aggregation and change map generation.
    \item \textbf{w/o 2DS}: This version removes the 2D Mamba Scan from both the encoder and decoder stages.  This ablation is designed to assess the contribution of the efficient long-range contextual modeling provided by the 2D Mamba Scan.
\end{itemize}

Table \ref{tab:ab} presents the results of these ablation studies.

\subsubsection{Impact of Change Flow Guidance}

Comparing the "Proposed" method with the "w/o ChangeFlow" variant across all datasets reveals the significant role of change flow guidance.  On WHU-CD, removing the flow guidance leads to a decrease of 3.73\% in Recall, 1.93\% in Precision, 0.28\% in OA, 2.87\% in F1-score, 5.06\% in IoU, and 3.01\% in KC.  Similar trends are observed on SYSU-CD, with reductions of 0.70\% in Recall, 5.88\% in Precision, 1.64\% in OA, 3.23\% in F1-score, 4.47\% in IoU, and 4.32\% in KC.  On LEVIR-CD+, the impact is also clear, with decreases of 2.15\% in Recall, 7.24\% in Precision, 0.42\% in OA, 4.62\% in F1-score, 7.04\% in IoU, and 4.83\% in KC. These consistent performance drops across all datasets when flow guidance is removed demonstrate that incorporating change flow significantly improves the model's ability to accurately identify change regions and maintain precision. The change flow information effectively guides the fusion of multi-temporal features and the subsequent decoding process, leading to more accurate change maps.

\subsubsection{Impact of 2D Mamba Scan}

The "w/o 2DS" variant, where the 2D Mamba Scan is removed, also shows performance degradation compared to the full model.  On WHU-CD, we observe a decrease of 2.21\% in Recall, a negligible change in Precision, a 0.12\% decrease in OA, a 1.15\% decrease in F1-score, a 2.05\% decrease in IoU, and a 1.2\% decrease in KC. On SYSU-CD, the removal of 2DS leads to a 1.84\% increase in Recall, a 3.23\% decrease in Precision, a 0.49\% decrease in OA, a 0.57\% decrease in F1-score, a 0.81\% decrease in IoU and a 0.92\% decrease in KC. For LEVIR+-CD, the removal of 2DS resulted in a 0.55\% decrease in Recall, a 1.76\% decrease in Precision, a 0.11\% decrease in OA, a 1.12\% decrease in F1-score, a 1.77\% decrease in IoU, and a 1.18\% decrease in KC. These results indicate that the 2D Mamba Scan plays a crucial role in capturing long-range contextual information, which is essential for accurate change detection. The consistent improvements observed across the three datasets confirm the importance of the 2DS module in enhancing the model's performance.

\begin{figure}[t]
    \centering
\includegraphics[width=0.98\linewidth]{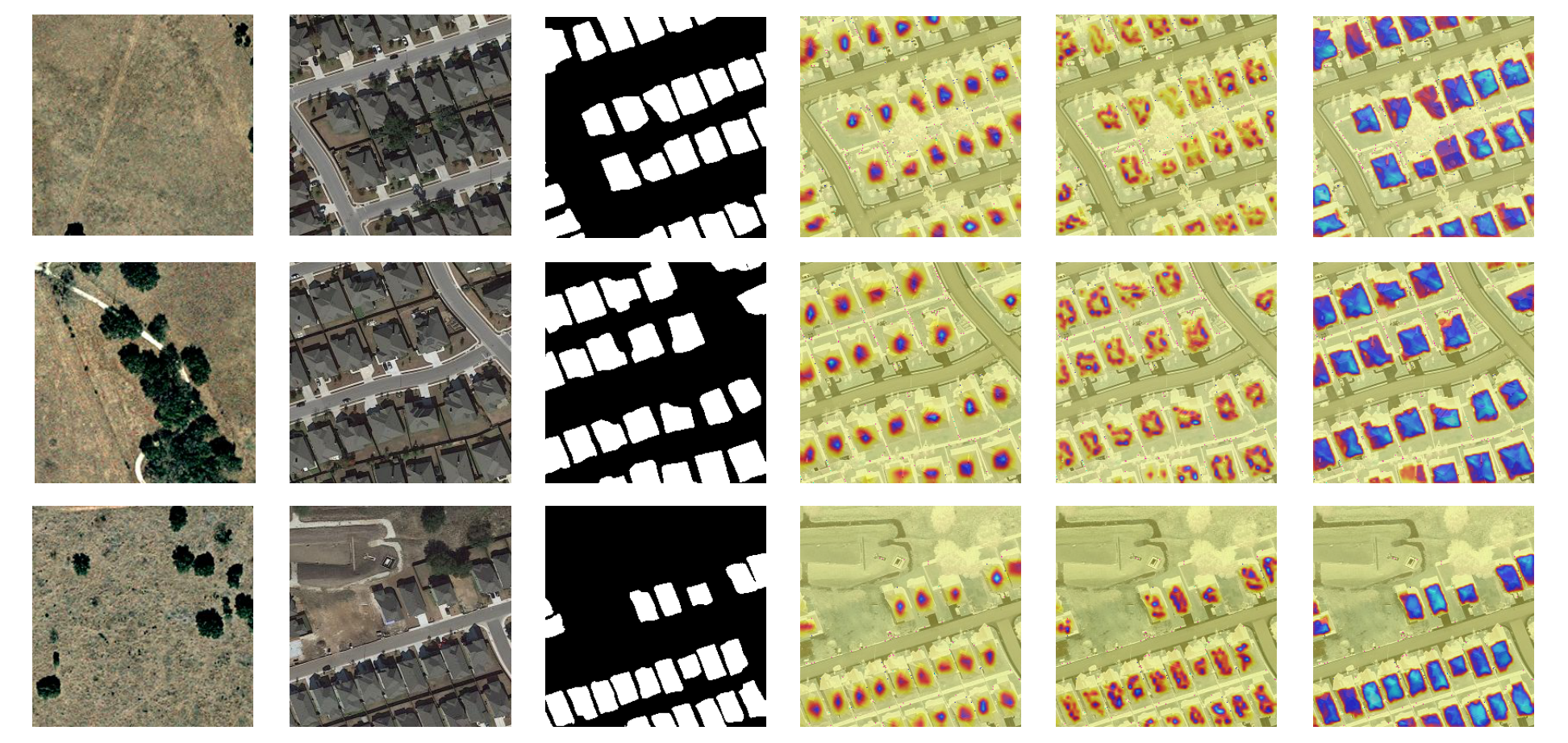}
    \caption{Layer-wise thermal activation maps of the proposed network. The figure showcases first column the pre-change image $I_1$, second column the post-change image $I_2$, third column the ground truth change mask and last three column the learned changemap. These visualizations highlight the network's focus on different aspects of change detection across its layers. Warmer colors (e.g., red, yellow) indicate higher activation, suggesting regions contributing significantly to the change detection process.}
    \label{fig:heat}
\end{figure}

\subsubsection{Summary of Ablation Study}

The ablation studies clearly demonstrate the individual contributions of both the change flow guidance and the 2D Mamba Scan to the overall performance of our proposed method.  Removing either component leads to a decrease in performance across all evaluation metrics and datasets, highlighting their complementary roles in achieving accurate change detection. The most significant performance drop is observed when the change flow guidance is removed, suggesting that it is a particularly critical component for precise change localization. The 2DS module, while also important, has a relatively smaller impact compared to the flow guidance, but still contributes significantly to performance gains.  These ablation studies provide strong evidence supporting the effectiveness of the proposed architecture and the importance of its constituent components.

\subsection{Visualization and Qualitative Analysis}

This section presents a qualitative analysis of the proposed method ("Ours") and compares its predictions with those of baseline models, as well as the ground truth. We also visualize the heatmaps of intermediate layers of our model to gain insights into its feature learning process.  In the comparative visualizations, false positives (FP) in the predictions are highlighted in \textcolor{red}{red}, while false negatives (FN) are highlighted in \textcolor{blue}{blue}.

\subsubsection{Comparative Visualization of Prediction Results}

\par Figures \ref{fig:whu}, \ref{fig:sysu}, and \ref{fig:levir+} provide a comprehensive visual comparison of change detection performance on the WHU, SYSU, and LEVIR+ datasets, respectively, between our proposed 2DMCG model, MambaCD, and HCGMNet. Each figure showcases pre- and post-change images, the corresponding ground truth annotations, and the change maps generated by each method.

\par On the WHU dataset (Figure \ref{fig:whu}), visual inspection reveals a clear advantage for our 2DMCG approach. The visual results, comprising pre- and post-change images, ground truth, and the outputs of all three methods, highlight the superior performance of our proposed 2DMCG technique. Our method effectively captures the complex change patterns present in the WHU dataset, whereas MambaCD and HCGMNet struggle with both commission and omission errors, particularly in areas with complex or subtle changes. 2DMCG's change flow guidance mechanism, derived from semantic flow, plays a crucial role in accurately decoding change information, leading to more refined and accurate change maps.

\par A similar analysis is presented for the SYSU dataset (Figure \ref{fig:sysu}). Our model demonstrates significantly better agreement with the ground truth, accurately identifying even subtle changes. In contrast, HCGMNet exhibits both false positives (commission errors) and misses (omission errors), indicating a lower level of precision and recall. MambaCD , while showing some success, also struggles with accurately delineating change boundaries, often producing fragmented or blurred change maps.  2DMCG's superior performance can be attributed to its ability to capture spatially continuous features using the 2D Mamba blocks, enabling a more precise representation of change regions.

\par The comparative results on the LEVIR+ dataset (Figure \ref{fig:levir+}) reinforce the observations from the previous datasets. The visual comparison, including pre- and post-change images, ground truth, and the respective change maps, further demonstrates the effectiveness of 2DMCG. Our method consistently aligns more closely with the ground truth, showcasing its ability to discern fine-grained changes even in complex urban environments. In contrast, both MambaCD and HCGMNet continue to exhibit limitations in accurately distinguishing changed and unchanged areas, yielding a higher rate of both false positives and false negatives.  The combination of 2D Mamba's spatial feature extraction and the change flow guided decoding process allows 2DMCG to outperform the competing methods across all datasets.

\begin{figure}[t]
    \centering
\includegraphics[width=0.98\linewidth]{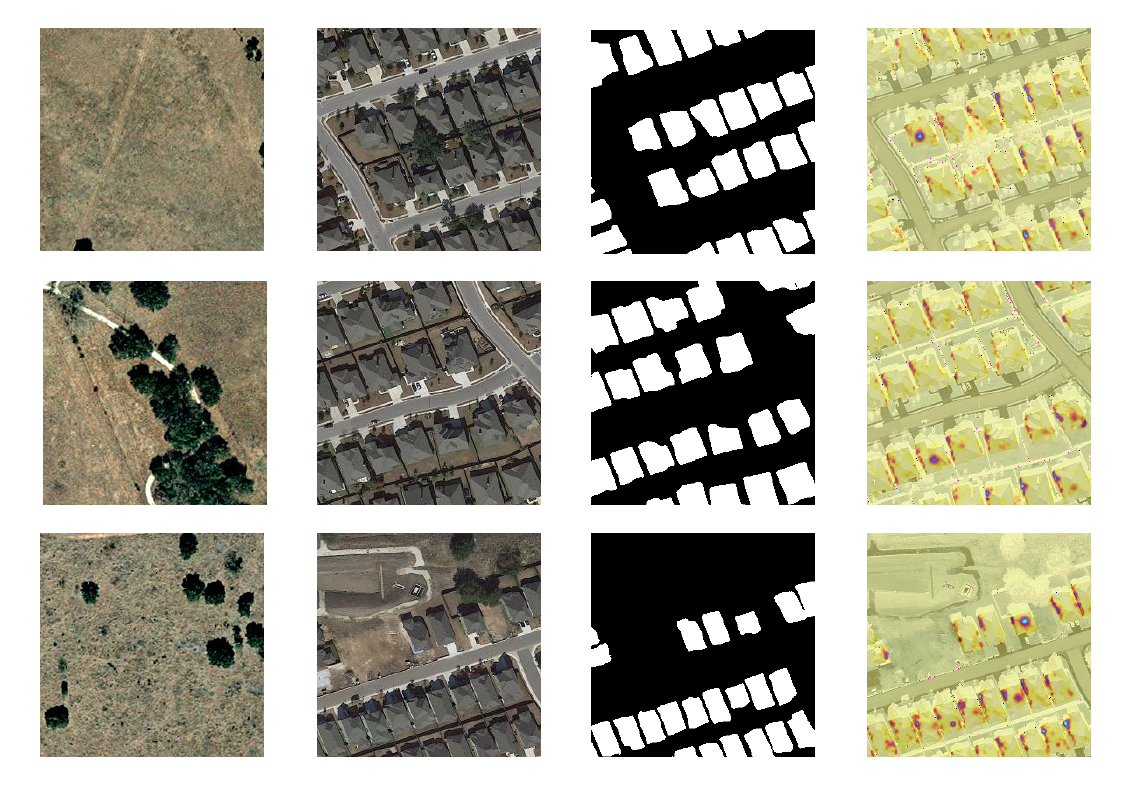}
    \caption{Visualization of ChangeFlow, illustrating its ability to capture feature correspondence analogous to optical flow.  The figure showcases first column the pre-change image $I_1$, second column the post-change image $I_2$, third column the ground truth change mask and fourth column the learned ChangeFlow field $\mathcal{F}$.  This flow field effectively warps features from $I_1$ towards $I_2$, facilitating direct comparison and accurate change detection. Warmer colors in the thermal activation maps indicate regions contributing significantly to the change detection process.}
    \label{fig:flow}
\end{figure}

\subsubsection{Heatmap Visualization of Intermediate Layers}

To better understand the feature learning process of our model, we visualize the heatmaps of the intermediate layers in Figure \ref{fig:heat}. These heatmaps illustrate the activation patterns of different neurons in the network, providing insights into which features are most discriminative for change detection. We observe that the earlier layers tend to capture low-level features such as edges and textures, while the deeper layers learn more complex and abstract features related to the semantic understanding of the scene and the changes within it. The heatmaps also show that our model focuses on the regions where changes have occurred.

Visualization of ChangeFlow thermal activation maps in Figure \ref{fig:flow}, inspired by optical flow techniques. Analogous to how optical flow captures apparent motion in video, ChangeFlow learns to model the "motion" or correspondence between features at different levels or instances.  These thermal maps visualize the network's focus on regions exhibiting significant feature change.  Warmer colors indicate higher activation, suggesting that these areas contribute most strongly to the learned "flow" and, consequently, to the change detection process.  Similar to semantic flow and feature warping, ChangeFlow leverages the concept of flow to enhance feature alignment and consistency for improved change detection.

\section{Conclusion}\label{sec:con}
This paper proposed an efficient framework based on a Vision Mamba variant to address challenges in remote sensing change detection (CD). While CNNs suffer from limited receptive fields and Transformers struggle with quadratic complexity, the Mamba architecture offers linear complexity and high parallelism. However, its 1D processing posed challenges in 2D vision tasks. Our framework enhances Mamba by effectively modeling 2D spatial information while maintaining its computational efficiency.
We introduced a 2DMamba encoder to capture global spatial context from multi-temporal images. For feature fusion, we used a 2D scan-based, channel-parallel scanning approach, combined with spatio-temporal fusion, effectively addressing spatial discontinuities. In the decoding phase, we proposed a change flow-based decoding method that improved feature map alignment.
Experiments on LEVIR-CD+ and WHU-CD demonstrated the superior performance of our framework over state-of-the-art methods, highlighting the potential of Vision Mamba for efficient and accurate CD in remote sensing.

\textbf{Limitations and Future Work:} Our framework assumes consistent image acquisition conditions, which limits robustness under varying illumination, scale, and rotation. Future work will focus on adapting the framework to handle such variations and optimizing the scanning patterns for specific change scenarios. We will also explore incorporating attention mechanisms within the 2DMamba block to enhance feature representation and capture long-range dependencies.

\printcredits

\bibliographystyle{cas-model2-names}

\bibliography{cas-refs}

\vskip3pt

\bio{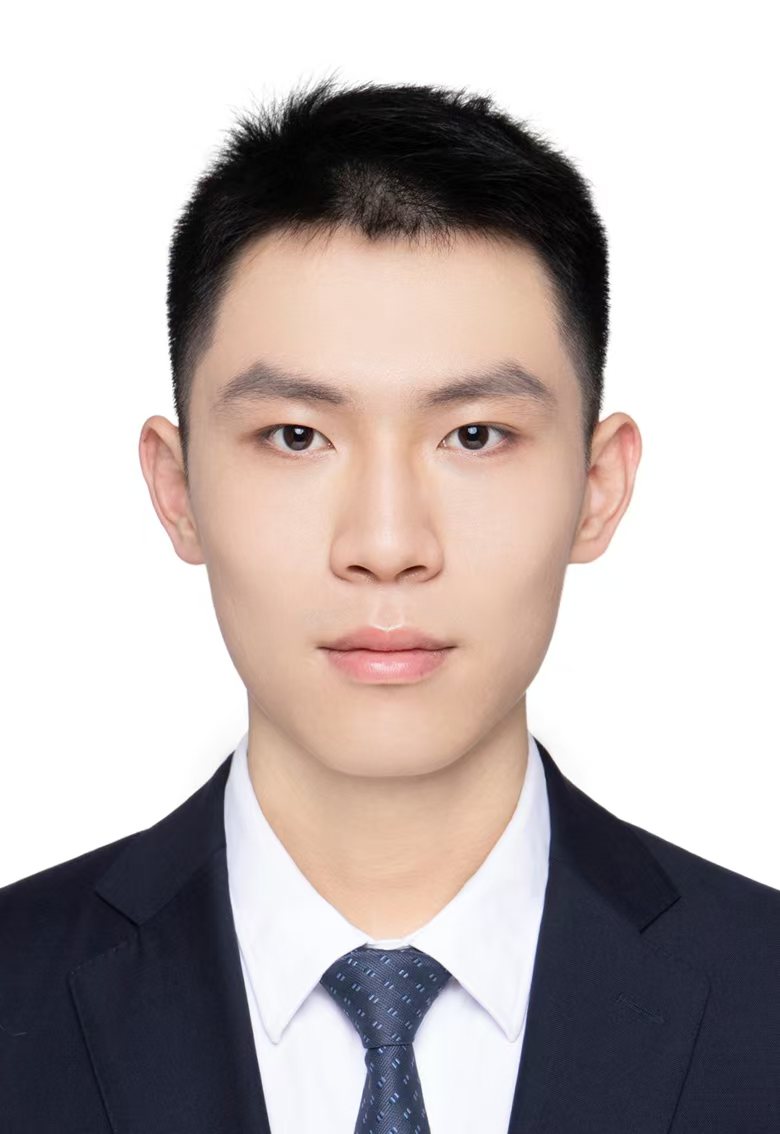}
Kuang Jun Yao was born in 2000. He is currently pursuing the M.S. degree in artificial intelligence at the School of Artificial Intelligence and Computer Science of Jiangnan University, China. His research interests include artificial intelligence, pattern recognition, image processing, and analysis.
\endbio

\vskip60pt

\bio{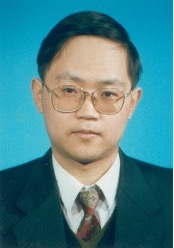}
Ge Hongwei was born in 1967. He received the M.S. degree in computer science from Nanjing University of Aeronautics and Astronautics, China, in 1992 and the Ph.D. degree in control engineering from Jiangnan University, China, in 2008. Currently, he is a professor and Ph.D. supervisor in the School of Artificial Intelligence and Computer Science of Jiangnan University. His research interests include artificial intelligence, pattern recognition, machine learning, image processing and analysis etc.
\endbio

\end{document}